\documentclass{article}

\PassOptionsToPackage{numbers, compress}{natbib}
 \usepackage[preprint]{neurips_2026}

\usepackage[utf8]{inputenc} % allow utf-8 input
\usepackage[T1]{fontenc}    % use 8-bit T1 fonts
\usepackage{hyperref}       % hyperlinks
\usepackage{url}            % simple URL typesetting
\usepackage{booktabs}       % professional-quality tables
\usepackage{amsfonts}       % blackboard math symbols
\usepackage{nicefrac}       % compact symbols for 1/2, etc.
\usepackage{microtype}      % microtypography
\usepackage{xcolor}         % colors
\usepackage{amsmath}
\usepackage{graphicx}
\graphicspath{{asset/appendix_viz/}}

\newcommand{\stackedfigure}[1]{%
  \begin{center}
    \includegraphics[width=\linewidth,height=0.43\textheight,keepaspectratio]{#1}
  \end{center}
  \vspace{1em}
}

\usepackage[normalem]{ulem}

\title{Loki: Representation over Architecture for Diffusion-Based Portrait Animation}

\author{%
  Pouyan Navard \\
  The Ohio State University \\
  \And
  Sernam Lim \\
  University of Central Florida \\
}

\begin{document}

\maketitle

\begin{figure}[h]
    \centering
    \includegraphics[width=\textwidth]{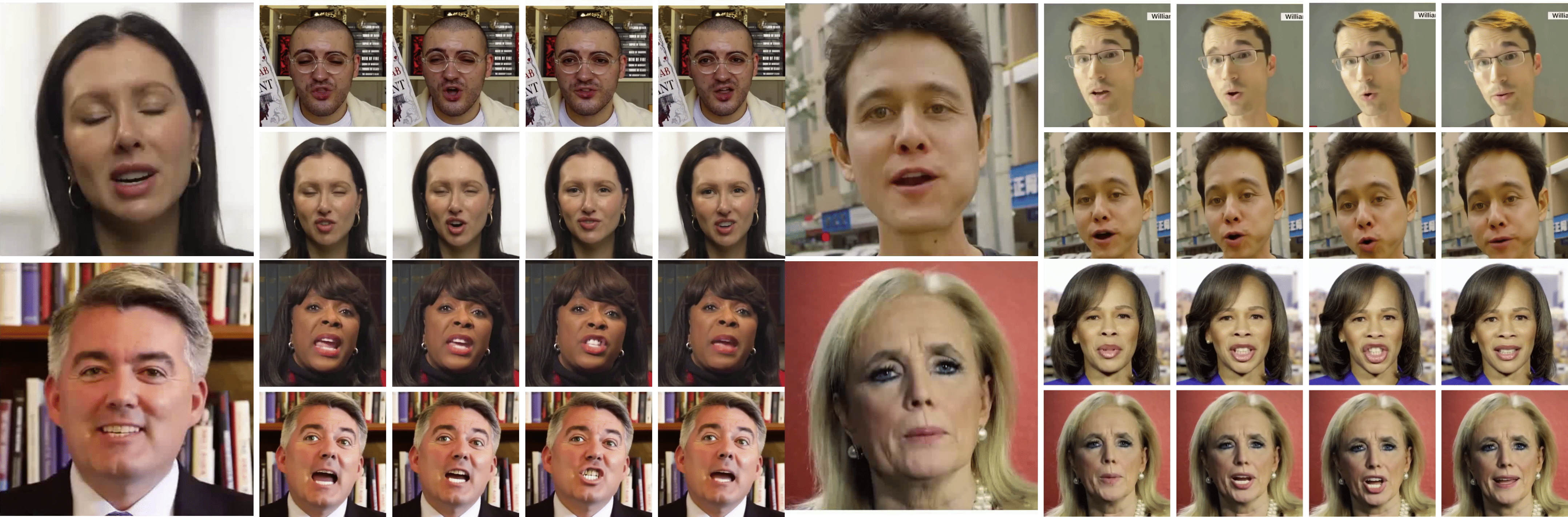}
    \caption{\textbf{Cross ID portrait animation with Loki.} Four examples, each showing a reference image (large, left) alongside four driver frames (odd rows) and the corresponding Loki outputs (even rows). Loki transfers the driver's expression and head pose onto the reference identity, preserving the reference's appearance while faithfully reproducing per-frame pose trajectory and fine-grained expression detail --- including large head rotations, wide mouth openings, and asymmetric expressions --- across diverse identity pairings.}
    \label{fig:teaser}
\end{figure}

\begin{abstract}
Portrait animation transfers a driver clip's facial expression and head pose onto a single reference image while preserving the reference's identity. State-of-the-art diffusion systems address this by stacking trained modules for expression, pose, and identity in turn, paying for it in trainable parameters, proprietary corpora, and residual entanglement between the very axes the system is meant to control independently. This complexity compensates for an upstream choice---learning facial expression and head pose from RGB, a representation in which identity, pose, and expression are inseparable without being learned apart. \textbf{Loki} steps out of RGB on the conditioning path. Driver expression and head pose are encoded by a face model whose parameter axes are identity-orthogonal by construction, then rasterised into a spatial map that the diffusion backbone consumes natively. Identity is routed separately through the diffusion backbone's own pretrained features via lightweight key-value injection. Because the parametric representation factorises identity from expression and pose, cross ID reenactment reduces to a coefficient substitution at inference, requiring no cross ID training data. Loki requires \textbf{$\sim$43\%} fewer inference parameters than leading diffusion baselines and trained on \textbf{1{,}496$\times$} less video samples. We define two metrics that directly measure whether the generated head pose trajectory and facial expression followed the driver's---the questions portrait animation actually asks; Loki leads or co-leads on both.

\end{abstract}

\section{Introduction}

Portrait animation transfers a driver clip's facial expression and head pose onto a single still reference image, frame by frame, while preserving the reference's identity~\cite{facevid2vid, emo, vexpress, aniportrait, liveportrait, hallo}. The setting of practical interest is \emph{cross ID}~\cite{marionette, huang2020learning, fsgan}, where the driver and reference are different people, as required for dubbing~\cite{dinet}, accessibility~\cite{pataranutaporn2021ai}, and virtual avatar applications~\cite{face2face, neuralvoicepuppetry}. Faithful results reduce to a single technical requirement: identity must be sourced from the reference and held constant across frames, while expression and head pose must be faithfully sourced from the driver~\cite{dpe}.

Recent diffusion-based systems address this by stacking trained modules dedicated to expression, pose, and identity~\cite{xportrait, hunyuanportrait, animateanyone, hallo, sadtalker, anitalker, echomimic}. Across this body of work, the conditioning signal is fundamentally RGB. This falls short: RGB entangles identity, pose, and expression at every pixel, so the disentanglement these modules perform is statistical rather than structural. The visible consequences---driver identity leaking into the output and fine-grained expression detail attenuated toward a generic prior~\cite{xportrait} and visible in our cross ID comparisons (Appendix~\ref{app:qualitative_examples}).

We present \textbf{Loki}, a portrait animation framework that conditions a pretrained diffusion backbone on FLAME~\cite{flame}, a parametric face model whose parameter groups for identity, expression, and pose are independent by construction. This disentangled representation is then rasterised into a spatial driver map that the diffusion backbone consumes natively, with two channel groups spatially defined on the face shape and invariant to identity. Together they tell the generation network two things at every pixel inside the face region: \emph{which point on the face is visible here}, and \emph{how has that point moved due to expression}. The first group gives each visible face point a unique coordinate at multiple spatial scales so that every point is distinguishable. The second captures how each point has displaced due to expression---a smile stretching lip corners, a jaw dropping, lips pushing forward in a pucker---and is naturally concentrated around the active regions of the face and near-zero everywhere else. Because the driver map carries no identity by construction, identity can be supplied entirely from the diffusion backbone's own rich visual representation, requiring no trained identity adapter.

Since expression and pose enter the generation process through the driver map and identity enters through a separate conditioning path, the two carry disjoint information by design. Cross ID reenactment follows naturally---at inference, the driver map is defined on the reference's face shape while carrying the driver's expression and pose, a parametric substitution that produces the same forward pass the model was trained on. No cross ID training data or dedicated reenactment module is needed. The result is a system that trains on under 2 hours of video with ${\sim}3.5\times$ fewer trainable parameters than leading diffusion baselines (Table~\ref{tab:resources}).

The evaluation contribution follows from the same parametric grounding. The metric suite the field has inherited from generic image and video generation~\cite{psnr, ssim, lpips, fvd, arcface} does not measure whether the prediction's head trajectory and facial expression followed the \emph{driver's}. We define two metrics directly on the parametric face model that fill this gap, and report results on both alongside the standard suite.

\paragraph{Contributions.}
\begin{itemize}
\item A portrait animation framework whose disentangled parametric conditioning removes the need for trained expression, pose, and identity modules---expression and pose are delivered through a single rasterised signal, and identity through the diffusion backbone's own generative prior---requiring ${\sim}$43\% fewer inference parameters than the leading diffusion baseline and 1{,}496$\times$ less training video (Table~\ref{tab:resources}).
\item A training recipe under which a model trained only on same-identity data performs cross ID portrait animation at inference by parametric substitution, with no cross ID training pairs.
\item Two evaluation metrics defined on the parametric face model that directly measure head pose and expression fidelity to the driver.

\end{itemize}

\begin{figure}[t]
\centering
\includegraphics[width=\linewidth]{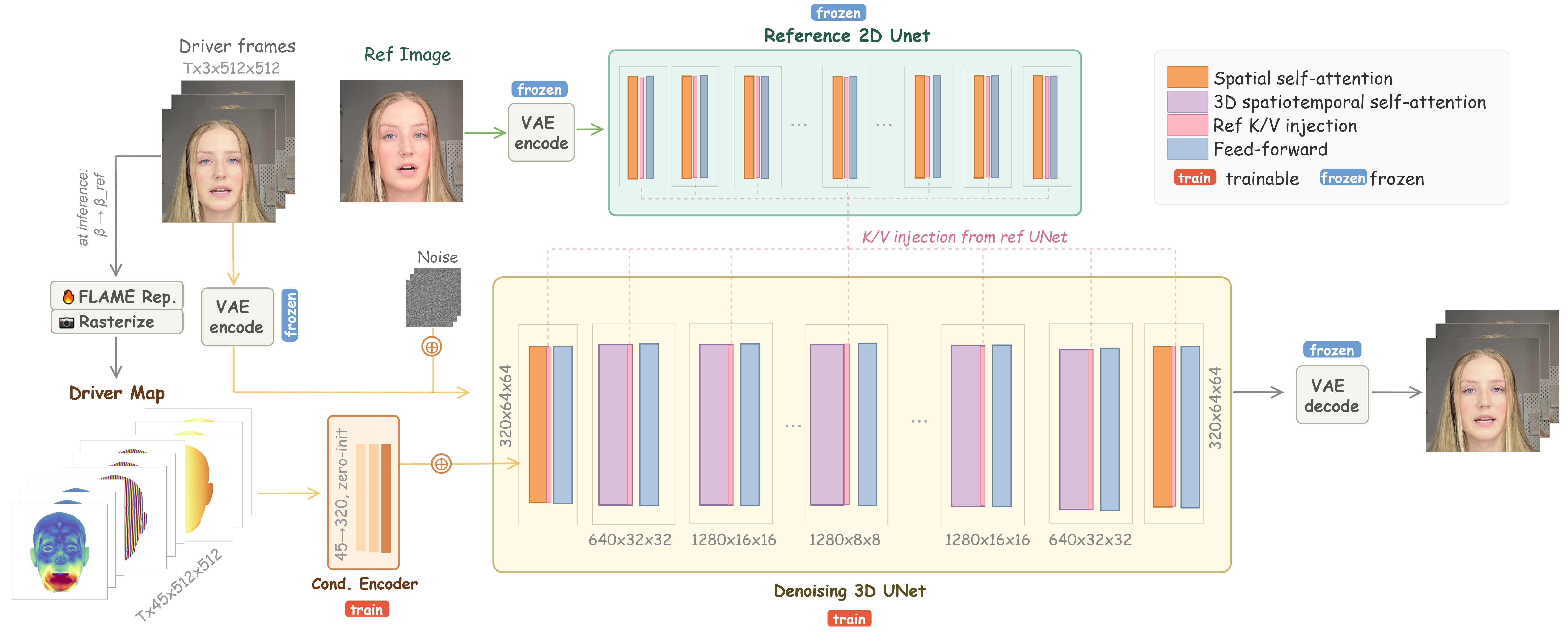}
\caption{\textbf{Loki pipeline overview.} The framework separates conditioning into two paths: a \emph{driver path} (left) that rasterises FLAME parameters into a spatial driver map encoding expression and pose, and an \emph{identity path} (top) that routes the reference image through a frozen copy of the pretrained 2D diffusion backbone via key-value injection. The driver map is downsampled by a lightweight zero-initialised convolutional encoder and added to the generation UNet's stem input features. The generation UNet (centre) is initialised from the same pretrained backbone and extended with spatiotemporal attention to denoise multiple frames jointly.}
\label{fig:architecture}
\end{figure}

\section{Related Work}

\paragraph{Portrait animation.}
Portrait animation has evolved from GAN-based systems that learn motion fields directly from RGB keypoints~\cite{fomm, facevid2vid} or add explicit disentanglement objectives to separate identity from expression~\cite{dpe, liveportrait}, to diffusion-based systems that replace the GAN generator with a denoising backbone, gaining image quality at the cost of additional conditioning infrastructure. The resulting architectures dedicate trained modules to each control axis: separate encoders for identity, pose, and expression~\cite{animateanyone, hallo}, multi-branch ControlNet~\cite{controlnet} conditioning for driver and reference features~\cite{xportrait}, or convolutional regressors paired with composite identity paths built from face recognition and self-supervised vision features~\cite{hunyuanportrait, ipadapter, arcface, dinov2}. Audio-driven methods follow the same template with audio replacing visual driving~\cite{sadtalker, anitalker, echomimic, emotalk, emote, voca}. A common thread runs through all of these: the driving signal originates in RGB or audio, and because neither representation excludes identity, preserving the reference's identity requires dedicated learned modules that must undo the entanglement at the input. Loki takes the opposite approach: identity is absent from the expression conditioning signal by construction, so no learned disentanglement is needed.

\paragraph{Parametric face models as conditioning.}
FLAME~\cite{flame} is a 3D morphable model that decomposes facial variation into independent parameter groups for identity shape, expression, and pose. It has been widely adopted for face reconstruction~\cite{deca, emoca, mica, spectre, smirk}, avatar generation~\cite{cap4d, gaussianavatars, flashavatar, imavatar, diffusionavatars}, and expression transfer~\cite{faceformer, codetalker}. Several portrait animation systems condition on FLAME or similar 3DMM parameters, but deliver them as global vectors rather than spatial signals: SadTalker~\cite{sadtalker} maps audio to 3DMM coefficients and injects them through a learned regressor, and FaceChain-ImagineID~\cite{facechainimagineid} projects expression coefficients into learned tokens consumed via cross-attention. Both routes discard the spatial inductive bias that makes a diffusion backbone effective---the model must learn \emph{where} on the face each coefficient applies, a mapping that is implicit in the parameterisation but absent from the conditioning signal. Loki rasterises FLAME into the backbone's native spatial domain, so spatial correspondence is given rather than learned, and we show empirically that this distinction accounts for a substantial performance gap (\S\ref{sec:exp-results}).

\paragraph{Evaluation.}
Portrait animation systems are predominantly evaluated with metrics inherited from generic image and video generation: PSNR, SSIM~\cite{ssim}, and LPIPS~\cite{lpips} for pixel fidelity, FVD~\cite{fvd} for distributional quality, and ArcFace cosine similarity~\cite{arcface} for identity preservation. We define two metrics grounded in the parametric face model that directly measure whether the generated head pose trajectory and facial expression followed the driver's.

\section{Method}
\label{sec:method}

\begin{figure}[t]
\centering
\includegraphics[width=\linewidth]{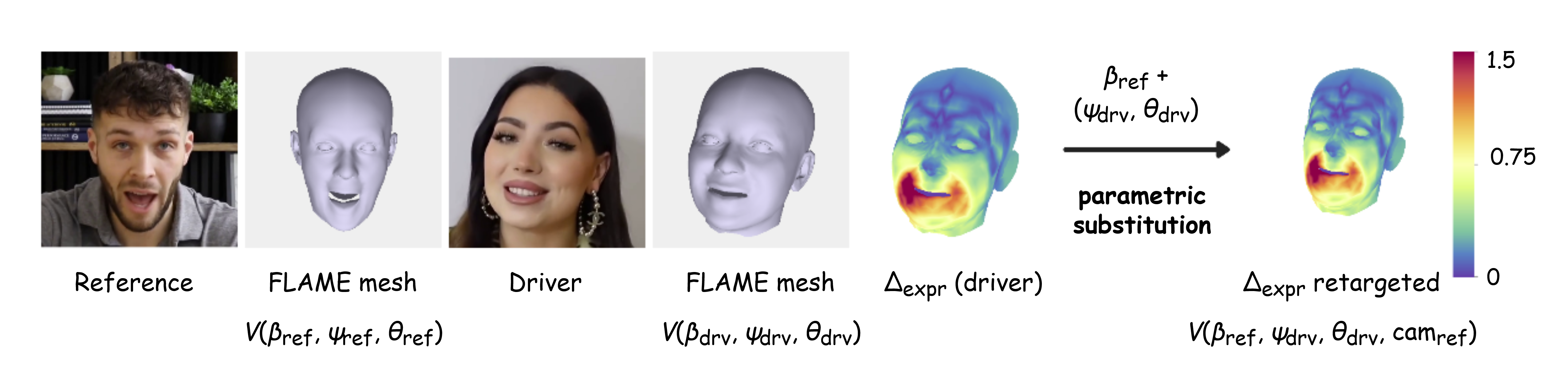}
\caption{\textbf{Parametric reenactment.} The reference's identity shape $\vec{\beta}_{\text{ref}}$ and camera are composed with the driver's expression $\vec{\psi}_{\text{drv}}$ and pose $\vec{\theta}_{\text{drv}}$, producing a driver map that carries the driver's expression on the reference's geometry. The substitution is a pure function with no learned parameters.}
\label{fig:retargeting}
\end{figure}

\subsection{Overview}
\label{sec:method-overview}
Loki feeds a single generation UNet from two strictly separate conditioning paths (Figure~\ref{fig:architecture}): a \emph{driver path} that encodes expression and head pose through a rasterised parametric face representation whose axes are identity-free by construction, and an \emph{identity path} that carries the reference's appearance through the diffusion backbone's own pretrained features. The two paths carry disjoint information by design; Section~\ref{sec:method-driver} and Section~\ref{sec:method-identity} describe each in turn.

\subsection{Driver Path}
\label{sec:method-driver}

\paragraph{Task-aligned face model.}
Portrait animation needs to control three axes independently: identity shape, facial expression, and head pose. FLAME~\cite{flame} is a 3D parametric face model whose parameter groups correspond exactly to these axes. It defines a fixed mesh topology of $N{=}5023$ vertices and a differentiable forward map:
\begin{equation}
\label{eq:flame}
M(\vec{\beta},\, \vec{\theta},\, \vec{\psi})
\;=\;
W\!\bigl(\,T_P(\vec{\beta},\, \vec{\theta},\, \vec{\psi}),\;
J(\vec{\beta}),\;
\vec{\theta},\;
\mathbf{W}\bigr),
\end{equation}
where $W$ denotes linear blend skinning~\cite{smpl} with blend weights $\mathbf{W}$ and joint locations $J(\vec{\beta})$. The posed template $T_P$ decomposes as
\begin{equation}
\label{eq:flame-tp}
T_P
\;=\;
\mathbf{T}
\;+\;
\underbrace{B_S(\vec{\beta};\,\mathbf{S})}_{\text{identity shape}}
\;+\;
\underbrace{B_P(\vec{\theta};\,\mathbf{P})}_{\text{pose correctives}}
\;+\;
\underbrace{B_E(\vec{\psi};\,\mathbf{E})}_{\text{expression}},
\end{equation}
where each offset field is a linear combination over an orthonormal basis: $B_S$ spans $|\vec{\beta}|{=}150$ identity coefficients, $B_E$ spans $|\vec{\psi}|{=}65$ expression coefficients, and $\vec{\theta}$ covers head, jaw, neck, and eye rotations. The property that makes FLAME suitable for portrait animation is that the three offset fields are \emph{independent}: the expression deformation $\Delta_{\text{expr}} = B_E(\vec{\psi};\,\mathbf{E})$ depends only on $\vec{\psi}$ and the fixed basis $\mathbf{E}$---it is unaffected by the identity coefficients $\vec{\beta}$ and by the pose $\vec{\theta}$. Symmetrically, changing $\vec{\psi}$ does not alter identity geometry, and changing $\vec{\theta}$ does not alter expression. This independence is the foundation of everything that follows: the driver map, the identity-free conditioning, and the cross ID capability at inference.

\paragraph{From parameters to pixels.}
FLAME provides the three control axes portrait animation needs, but its native representation is a parameter vector---a flat, non-spatial signal. As discussed in \S2, prior work that conditions on 3DMM parameters operates in this vector space, discarding the spatial inductive bias that makes a diffusion UNet effective. We show empirically in \S\ref{sec:exp-results} that this distinction accounts for a substantial performance gap.

\begin{figure}[t]
\centering
\includegraphics[width=\linewidth]{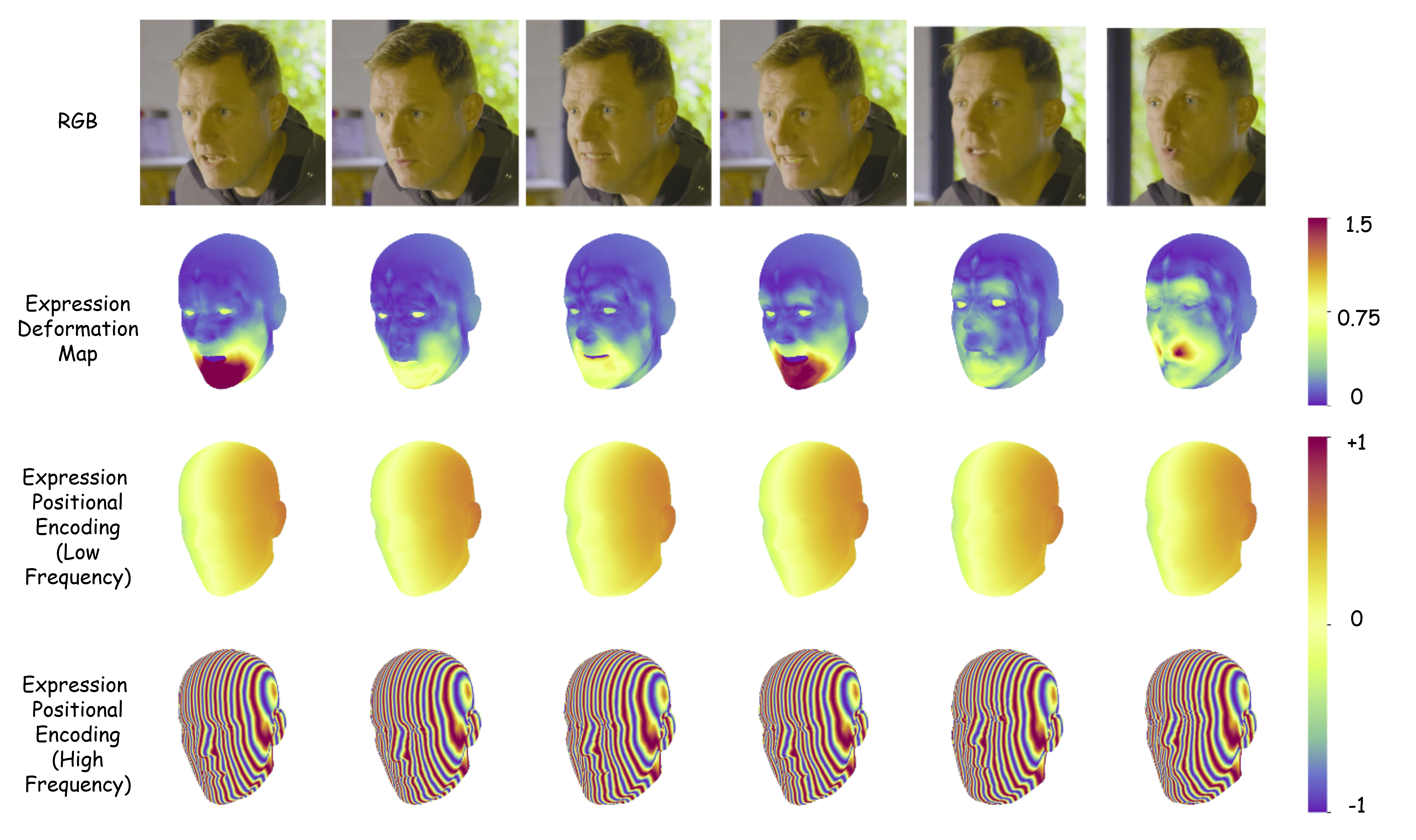}
\caption{\textbf{The driver map.} Six frames from a driver clip shown across columns. From top to bottom: (1)~the input RGB frame; (2)~the expression deformation magnitude $\|\Delta_{\text{expr}}\|_2$ (colorbar shared across frames; near-zero wherever the face has not moved off its neutral expression); (3--4)~a low-frequency ($\sin(2^0 x)$) and high-frequency ($\sin(2^6 x)$) slice of the positional encoding---two of the 42 channels; the full set is visualised in Appendix~\ref{app:driver-map-channels}. Rows 2--4 together compose the $45$-channel driver map: 3 expression deformation channels and 42 positional encoding channels.}
\label{fig:conditioning-map}
\end{figure}

Instead, we rasterise the FLAME representation into a spatial map at the backbone's input resolution ($512{\times}512$), which we call the \emph{driver map} (Figure~\ref{fig:conditioning-map}). The driver map decomposes into two channel groups that preserve FLAME's axis independence through rasterisation; the full channel specification is given in Appendix~\ref{app:driver-map-channels}.

\paragraph{A coordinate system for the face.}
The first group identifies \emph{which} canonical face point is visible at each pixel. We take FLAME's template-space vertex positions and apply a sinusoidal positional encoding (Figure~\ref{fig:conditioning-map}, rows 3--4):

\begin{equation}
\label{eq:posenc}
\gamma(\mathbf{p})
\;=\;
\bigl[\,
\sin(2^0 \mathbf{p}),\;
\cos(2^0 \mathbf{p}),\;
\ldots,\;
\sin(2^6 \mathbf{p}),\;
\cos(2^6 \mathbf{p})
\,\bigr]
\;\in\; \mathbb{R}^{42},
\end{equation}
giving $3 \text{ axes} \times 7 \text{ octaves} \times 2 \text{ (sin/cos)} = 42$ channels. These are rasterised through the FLAME mesh into the $512{\times}512$ image. Because the encoded values are template-space coordinates, the same canonical point always produces the same encoding regardless of identity---the live per-frame vertices influence the encoding only indirectly, by determining which template point is visible at each pixel through the rasterisation geometry. Low-frequency bands carry smooth spatial gradients across the face, while high-frequency bands produce fine fringes that give each point a unique signature.

\paragraph{A displacement signal for expression.}
The second group captures \emph{how} each visible point has moved due to expression. Each FLAME vertex carries a 3D expression offset $\Delta_{\text{expr}} = (\Delta x, \Delta y, \Delta z)$---the displacement from the neutral face to the current expression. These per-vertex offsets are rasterised into the $512{\times}512$ image, so every on-mesh pixel inherits the deformation of the visible vertex at that location. The three channels carry direct geometric meaning: $\Delta x$ captures lateral displacement (lip corners stretching in a smile), $\Delta y$ captures vertical displacement (jaw drop, brow raise), and $\Delta z$ captures depth displacement (lips puckering forward, cheeks puffing). Because $\Delta_{\text{expr}} = B_E(\vec{\psi};\,\mathbf{E})$ depends only on $\vec{\psi}$ (Eq.~\ref{eq:flame-tp}), the deformation is zero for a neutral expression regardless of identity or pose, and is naturally concentrated around the active regions of the face.

\begin{figure}[h]
\centering
\includegraphics[width=\textwidth]{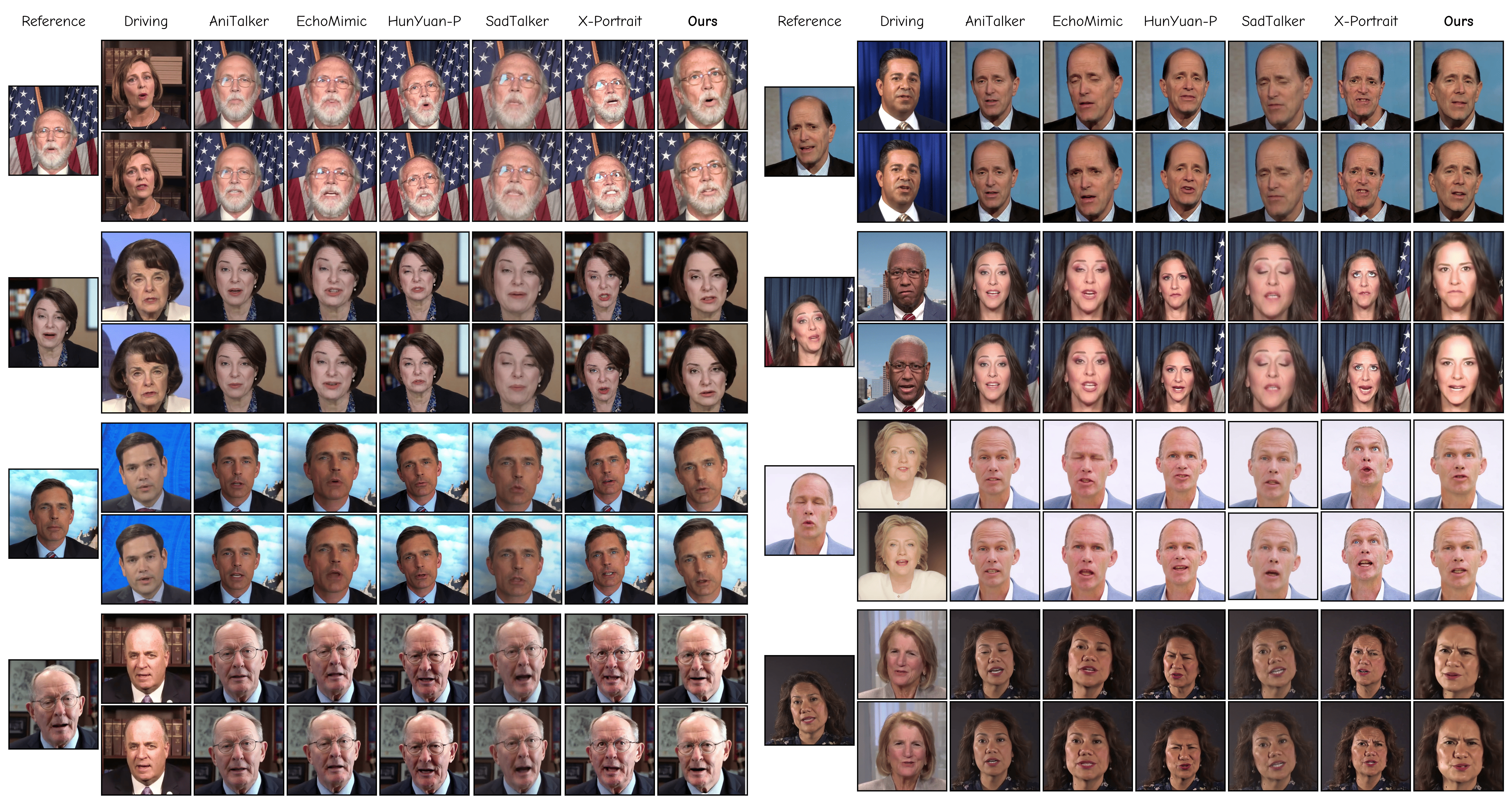}
\caption{\textbf{Qualitative comparison on cross ID portrait animation (HDTF).} The figure is organized into two panels, each containing four example pairs. Within each pair, a single reference image (left column) is animated using two different driving frames (one per row). Columns show outputs from each baseline and our method. Baselines either under-rotate the head (SadTalker, AniTalker, EchoMimic), attenuate expression amplitude toward a moderate prior (HunyuanPortrait), or introduce expression artifacts that distort the driver's facial configuration (X-Portrait). Loki reproduces both the driver's head orientation and expression intensity with higher fidelity per frame. Please zoom in.}
\label{fig:qualitative}
\end{figure}

\paragraph{From same-identity training to cross ID inference.}
The independence established in Eq.~\ref{eq:flame-tp} has a direct practical consequence. Consider what happens when we evaluate FLAME with the reference's identity shape $\vec{\beta}_{\text{ref}}$ but the driver's expression $\vec{\psi}_{\text{drv}}$ and pose $\vec{\theta}_{\text{drv}}$:
\begin{equation}
\label{eq:retarget}
V(\vec{\beta}_{\text{ref}},\, \vec{\psi}_{\text{drv}},\, \vec{\theta}_{\text{drv}},\, \text{cam}_{\text{ref}}).
\end{equation}
The positional encoding channels are unchanged---they encode template-space coordinates that do not depend on $\vec{\beta}$. The expression deformation channels are unchanged---$\Delta_{\text{expr}} = B_E(\vec{\psi}_{\text{drv}};\,\mathbf{E})$ depends only on $\vec{\psi}_{\text{drv}}$. What changes is only the spatial layout: the rasterisation projects onto the reference's face geometry through the reference's camera, so the driver map aligns with the reference's pixel frame. The generation UNet therefore sees a driver map that is identical in content to what it encountered during same-identity training but laid out on a different face shape (Figure~\ref{fig:retargeting}). Cross ID reenactment is a property of the conditioning representation, not a learned capability.

\subsection{Identity Path}
\label{sec:method-identity}

Because the driver map carries no identity by construction, the identity path needs to carry only the reference's appearance---and can do so without a dedicated trained module. A frozen copy of the Stable Diffusion 2.1 UNet, initialised from the same checkpoint as the generation UNet, processes the VAE-encoded reference image once and exports its per-layer self-attention features into the generation UNet's matching layers as additional keys and values:
\begin{equation}
K = W_K [\mathbf{x}_{\text{gen}};\, \mathbf{x}_{\text{ref}}], \quad V = W_V [\mathbf{x}_{\text{gen}};\, \mathbf{x}_{\text{ref}}],
\end{equation}
where $W_K$ and $W_V$ are the generation UNet's \emph{own} projection matrices. This works without a learned adapter because both UNets share the same architecture and initialisation---the reference features already live in the generation UNet's native representation space. No auxiliary identity loss is applied; the standard denoising objective alone is sufficient to learn the identity routing. Implementation details are given in Appendix~\ref{app:identity-path}.

\subsection{Generation Backbone}
\label{sec:method-backbone}

\paragraph{Backbone.}
Loki operates in the latent space of a frozen Stable Diffusion 2.1~\cite{stablediffusion} VAE, which encodes a $512{\times}512$ RGB frame to a $64{\times}64{\times}4$ latent. The generation UNet is trained under the standard $\varepsilon$-prediction objective (Appendix~\ref{app:noise-schedule}); at inference, clean latents are recovered by DDIM sampling~\cite{ddim} with $50$ steps and classifier-free guidance~\cite{cfg} at scale $s{=}2.0$. The generation UNet inherits the spatial hierarchy of Stable Diffusion 2.1 ($64^2 \to 32^2 \to 16^2 \to 8^2$), with 3D spatiotemporal attention replacing per-frame self-attention at the three inner resolutions to enforce temporal coherence (Figure~\ref{fig:architecture}); per-frame spatial attention is retained at $64^2$ where the full spatiotemporal sequence is prohibitively long. Both modes consume reference key-value tokens from the identity path (\S\ref{sec:method-identity}). The driver map is projected into the UNet's feature space by a lightweight convolutional encoder whose residual blocks mirror the UNet's own style, with a zero-initialised output convolution~\cite{controlnet} that ensures training starts from the pretrained state. Token-layout and encoder details are given in Appendices~\ref{app:attention-shapes} and~\ref{app:cond-encoder}.

\paragraph{Conditioning encoder.}
The driver map is a $45$-channel tensor at $512{\times}512$ that must be projected into the generation UNet's $64{\times}64$ feature space. A lightweight convolutional encoder performs this projection, using residual blocks that mirror the generation UNet's own convolutional style so that the features it produces are compatible with the UNet's input expectations. The output convolution is zero-initialised so the generation UNet begins training from its pretrained state, with the conditioning signal learned gradually~\cite{controlnet}. The output is added element-wise to the generation UNet's post-stem feature map. The full specification is given in Appendix~\ref{app:cond-encoder}.

\begin{table}[t]
\centering
\caption{\textbf{Resource comparison across methods.} Training corpus size (upper-bound clip count and total video duration) and total inference pipeline parameters, with ratios relative to Loki. Loki trains on a fraction of the data used by all baselines while maintaining a compact inference pipeline relative to diffusion-based systems.}
\label{tab:resources}
\small
\setlength{\tabcolsep}{4pt}
\begin{tabular}{l cc cc cc}
\toprule
Method & Clips & $\times$Loki & Duration (${\sim}$hr) & $\times$Loki & Params & $\times$Loki \\
\midrule
SadTalker~\cite{sadtalker}       & 100{,}000  & 9.4$\times$ & 352    & 207$\times$ & 273M & 0.15$\times$ \\
AniTalker~\cite{anitalker}       & 17{,}108   & 1.6$\times$ & 55     & 32$\times$  & 537M & 0.30$\times$ \\
EchoMimic~\cite{echomimic}       & 166{,}000  & 15.6$\times$ & 621   & 365$\times$ & 2.24B & 1.27$\times$ \\
X-Portrait~\cite{xportrait}      & 23{,}650   & 2.2$\times$ & ---    & ---         & 3.11B & \textbf{1.76}$\times$ \\
HunyuanPortrait~\cite{hunyuanportrait} & 202{,}500 & \textbf{19.0}$\times$ & 2{,}543 & \textbf{1{,}496$\times$} & 2.18B & 1.23$\times$ \\
\midrule
\textbf{Loki}            & 10{,}649   & 1$\times$ & 1.7    & 1$\times$   & 1.77B & 1$\times$ \\
\bottomrule
\end{tabular}
\end{table}

\section{Experiments}
\label{sec:experiments}

\subsection{Setup}
\label{sec:exp-setup}

\paragraph{Datasets.}
Loki is trained on a subset of TalkVid~\cite{talkvid}, comprising 10{,}649 clips, each containing 16 frames at 25\,fps, from 1{,}576 distinct identities (1{,}493 training, 83 validation for hyperparameter tuning only). The identity distribution is long-tailed: approximately 75\% of identities appear in two clips and 18\% in only one. Each clip is face-cropped using ArcFace~\cite{arcface} and resized to $512{\times}512$. All results are evaluated on HDTF~\cite{hdtf}, a fully held-out dataset of 212 identities with one clip each, under two protocols: \emph{same-identity reconstruction} and \emph{cross ID reenactment}.

\paragraph{Baselines.}
We compare against five systems spanning visual-driven and audio-driven paradigms: X-Portrait~\cite{xportrait} and HunyuanPortrait~\cite{hunyuanportrait} (visual-driven), and SadTalker~\cite{sadtalker}, AniTalker~\cite{anitalker}, and EchoMimic~\cite{echomimic} (audio-driven). All baselines use official code and released checkpoints; audio-driven methods receive the driver clip's audio as input. Table~\ref{tab:resources} compares training corpus size and inference pipeline parameters across all methods. Loki trains on roughly 10k clips totalling under 2 hours of video---between one and three orders of magnitude less than every baseline---while its inference pipeline (1.77B parameters) is the most compact among diffusion-based systems.

\subsection{Training and Inference}
\label{sec:exp-training}

\paragraph{Training.}
Each training sample is drawn from a single clip of one identity. A window of $T{=}16$ consecutive frames forms the denoising target, and a reference frame is sampled from a different temporal position in the same clip---deliberately outside the target window so that the identity path is forced to carry only what is identity-invariant while the driver path carries only what varies with expression and pose. Classifier-free guidance dropout at $p{=}0.1$ zero-fills the entire conditioning set. Only the generation UNet (${\sim}823$M parameters) and the conditioning encoder (${\sim}7.35$M) receive gradients; all other components are frozen. Training runs for $30$k steps with batch size $4$ per GPU across $8{\times}$H200 GPUs, using AdamW at learning rate $10^{-4}$.

\paragraph{Inference.}
The forward pass is identical to training except that FLAME parameters are either used directly (same-identity) or composed as $V(\vec{\beta}_{\text{ref}}, \vec{\psi}_{\text{drv}}, \vec{\theta}_{\text{drv}}, \text{cam}_{\text{ref}})$ for cross ID (\S\ref{sec:method-driver}), and denoising uses DDIM with $50$ steps at guidance scale $s{=}2.0$.

\subsection{Evaluation Metrics}
\label{sec:exp-metrics}

The metric suite inherited from generic image and video generation---PSNR, SSIM~\cite{ssim}, LPIPS~\cite{lpips}, FVD~\cite{fvd}, ArcFace cosine similarity~\cite{arcface}---does not directly measure whether a portrait animation system reproduced the driver's head trajectory and facial expression. We report these for comparability but introduce two metrics defined on the parametric face model that measure these questions directly. Both are computed by fitting FLAME to every predicted and ground-truth frame under identical conditions using the same off-the-shelf reconstruction pipeline.

\paragraph{Head Pose Follow (HPF).}
HPF measures how faithfully the generated head orientation follows the driver's rotation trajectory. Because predicted and target clips are reconstructed independently with separate cameras, absolute orientations are not comparable. We therefore compare frame-0-anchored delta rotations:
\begin{equation}
\label{eq:delta-rot}
\Delta R[t] = R_{\text{head}}[t] \cdot R_{\text{head}}[0]^\top,
\end{equation}
which cancels the per-clip camera offset and isolates the pose trajectory. The per-frame error is the geodesic angular distance between predicted and target deltas, reported in degrees and averaged over the evaluation window. Full derivation is given in Appendix~\ref{app:hpf}.

\begin{table}[t]
\centering
\caption{\textbf{Comparison on HDTF.} cross ID reenactment (left) and same-identity reconstruction (right). HPF (Head Pose Follow) and HEF (Expression Follow) are defined in \S\ref{sec:exp-metrics}. PSNR, SSIM, LPIPS, FVD, and ID-Cos (ArcFace cosine similarity) are included for comparability with prior work. Best in \textbf{bold}, second-best \underline{underlined}.}
\label{tab:results}
\small
\setlength{\tabcolsep}{3pt}
\begin{tabular}{l cc c cc cccc}
\toprule
 & \multicolumn{3}{c}{\emph{Cross ID}} & \multicolumn{6}{c}{\emph{Same ID}} \\
\cmidrule(lr){2-4} \cmidrule(lr){5-10}
Method & HPF$\downarrow$ & HEF$\downarrow$ & ID-Cos$\uparrow$ & HPF$\downarrow$ & HEF$\downarrow$ & PSNR$\uparrow$ & SSIM$\uparrow$ & LPIPS$\downarrow$ & FVD$\downarrow$ \\
\midrule
SadTalker       & 3.217 & 0.1107 & \textbf{0.921} & 3.423 & 0.0811 & 14.917 & 0.609 & 0.453 & 78.436 \\
AniTalker       & 3.031 & 0.1134 & 0.815 & 3.101 & 0.0963 & 17.390 & 0.687 & 0.306 & 102.635 \\
EchoMimic       & 3.268 & 0.1156 & \underline{0.897} & 3.307 & 0.0823 & 14.779 & 0.595 & 0.431 & 76.422 \\
X-Portrait      & 2.945 & 0.1053 & 0.825 & 2.145 & \textbf{0.0652} & \textbf{25.483} & \textbf{0.844} & \textbf{0.091} & \textbf{47.348} \\
HunyuanPortrait & \textbf{2.193} & \underline{0.0850} & 0.823 & \textbf{2.076} & \underline{0.0658} & \underline{23.213} & \underline{0.793} & \underline{0.119} & \underline{56.361} \\
\midrule
Loki      & \underline{2.243} & \textbf{0.0828} & 0.791 & \underline{2.134} & 0.0664 & 14.235 & 0.588 & 0.435 & 163.108 \\
\bottomrule
\end{tabular}
\end{table}

\paragraph{Expression Follow (HEF).}
HEF measures how faithfully the generated facial expression matches the driver's, isolating expression from head pose, identity shape, and camera. At each frame, the prediction's expression parameters are inserted into the target's full FLAME fit (retaining the target's pose, shape, and camera), and two deformation maps are rendered---one from the target's original fit, one from the substituted fit:
\begin{equation}
\label{eq:expr-sub}
\Delta_{\text{expr}}^{\text{sub}} = \text{rasterise}\!\bigl(\,
V(\vec{\beta}_{\text{tgt}},\; \vec{\psi}_{\text{pred}},\; \vec{\theta}_{\text{tgt}},\; \text{cam}_{\text{tgt}})\bigr).
\end{equation}
Because only expression differs, both maps occupy the same pixels and any residual is attributable purely to expression mismatch. Full derivation is given in Appendix~\ref{app:hef}.

\subsection{Results}
\label{sec:exp-results}

\paragraph{Quantitative comparison.}
Table~\ref{tab:results} reports results on HDTF. On cross ID reenactment, Loki leads on HEF ($0.083$ vs.\ $0.085$ for HunyuanPortrait) and is a close second on HPF ($2.24°$ vs.\ $2.19°$). To contextualise HEF: ${\sim}0.03$ is the practical floor, ${\sim}0.10$ is the no-skill baseline, and ${\sim}0.18$ is the realistic ceiling (Appendix~\ref{app:hef-calibration}). Under this calibration, Loki and HunyuanPortrait fall in the strong-match range, while the audio-driven cluster at $0.11$--$0.12$ is at or above the no-skill baseline.

\begin{table}[t]
\centering
\caption{\textbf{Ablation study on HDTF.} Each variant modifies only the conditioning representation; architecture and training are identical. \textit{Raw vector}: same FLAME parameters delivered as a projected vector (no rasterisation). \textit{w/o deformation}: the 3 expression deformation channels removed from the driver map. \textit{w/o posenc}: the 42 positional encoding channels removed from the driver map.}
\label{tab:ablation}
\small
\setlength{\tabcolsep}{3pt}
\begin{tabular}{l cc c cc ccc}
\toprule
 & \multicolumn{3}{c}{\emph{Cross ID}} & \multicolumn{5}{c}{\emph{Same-identity}} \\
\cmidrule(lr){2-4} \cmidrule(lr){5-9}
Variant & HPF$\downarrow$ & HEF$\downarrow$ & ID-Cos$\uparrow$ & HPF$\downarrow$ & HEF$\downarrow$ & PSNR$\uparrow$ & SSIM$\uparrow$ & LPIPS$\downarrow$ \\
\midrule
Loki (full)    & 2.243 & 0.0828 & 0.791 & 2.134 & 0.0664 & 14.235 & 0.588 & 0.435 \\
\midrule
\textit{raw vector}        & 3.435 & 0.1167 & 0.883 & 3.668 & 0.0933 & 14.023 & 0.567 & 0.468 \\
\textit{w/o deformation}   & 3.435 & 0.1435 & 0.728 & 3.668 & 0.1213 & 13.560 & 0.507 & 0.500 \\
\textit{w/o posenc}        & 4.124   & 0.0987    & 0.760 & 2.196 & 0.0654 & 13.609 & 0.544 & 0.473 \\
\bottomrule
\end{tabular}
\end{table}

The inherited metrics produce a ranking that diverges from HPF and HEF. X-Portrait leads on pixel-aligned metrics (PSNR, SSIM, LPIPS) yet ranks third on both HPF and HEF; SadTalker leads on ID-Cos ($0.921$) yet ranks last on both. Loki ranks last on every inherited metric yet first or second on HPF and HEF. The divergence reflects what each group measures: pixel-aligned metrics and FVD reward systems whose conditioning preserves the ground-truth framing, while ID-Cos penalises the pose and expression variation that faithful driver-following produces.

\paragraph{Qualitative comparison.}
Figure~\ref{fig:qualitative} shows representative cross ID results. Audio-driven methods under-rotate the head and dampen expression relative to the driver. Among visual-driven baselines, HunyuanPortrait follows head pose but attenuates expression amplitude---wide mouth openings and raised brows are softened toward a moderate prior---while X-Portrait occasionally introduces expression artifacts. Loki reproduces both the driver's head orientation and expression intensity with higher per-frame fidelity. Additional examples are given in Appendix~\ref{app:qualitative_examples}.

\paragraph{Ablations.}
Table~\ref{tab:ablation} isolates the contribution of each component in the driver map. The \textit{raw vector} variant receives exactly the same FLAME parameters as the full model but delivers them as a projected vector spatially broadcast to $64{\times}64$ following the technique used in~\cite{spbroaddec, film}, bypassing rasterisation--the same conditioning strategy used by SadTalker~\cite{sadtalker} and FaceChain-ImagineID~\cite{facechainimagineid}. HPF degrades from $2.13°$ to $3.67°$ on same-identity ($+72\%$) and HEF from $0.066$ to $0.093$ ($+41\%$), falling to the level of audio-driven baselines despite having access to the same parametric signal. The UNet receives the right information in the wrong form---without spatial structure, it must learn where on the face each coefficient applies, and on the same training budget it cannot recover what rasterisation provides for free.

Removing the 3 deformation channels (\textit{w/o deformation}) degrades HEF most severely---from $0.066$ to $0.121$ on same-identity ($+83\%$)---confirming that expression fidelity flows through the deformation signal. Removing the 42 positional encoding channels (\textit{w/o posenc}) degrades cross ID HPF most severely ($4.12°$, the worst of all ablations) and increases cross ID HEF by $19\%$, while same-identity expression fidelity remains largely intact. This asymmetry is revealing: without the coordinate system, the model can still handle same-identity inputs---where the spatial layout of the driver map matches what was seen during training---but fails at cross ID, where the driver map is laid out on an unfamiliar face shape and the positional encoding is needed to anchor the spatial correspondence. Both ablations support the driver map's two-group design.

\section{Conclusion}

Loki demonstrates that the architectural complexity of recent portrait animation systems can be replaced by a representational choice: conditioning a diffusion backbone on a parametric face model whose axes are identity-free by construction. This yields a system requiring \textbf{${\sim}$43\%} fewer inference parameters and \textbf{1{,}496$\times$} less training video than the leading diffusion baseline, in which cross-identity reenactment emerges as a parametric substitution with no cross-identity data or dedicated retargeting. The same parametric grounding motivates two evaluation metrics that expose a systematic divergence between inherited metrics and faithful driver-following. Loki's conditioning is bounded by the fidelity of the underlying FLAME fits, and the pixel-metric gap relative to RGB-conditioned baselines remains open; bridging it without reintroducing identity entanglement is a direction for future work.

%%%%%%%%%%%%%%%%%%%%%%%%%%%%%%%%%%%%%%%%%%%%%%%%%%%%%%%%%%%%
\newpage
\appendix
 
%% =============================================================
%% APPENDIX A — ADDITIONAL RESULTS
%% =============================================================
\section{Additional Results}
\label{app:qualitative_examples}
 
This appendix presents additional generation results beyond the examples shown in the main paper.
 
\subsection{Cross-Identity Examples}
\label{app:cross ID-examples}

\stackedfigure{cross_identity/hdtf__cross_identity__id_0040_id_0157.png}
\stackedfigure{cross_identity/hdtf__cross_identity__id_0013_id_0166.png}
\stackedfigure{cross_identity/hdtf__cross_identity__id_0025_id_0202.png}
\stackedfigure{cross_identity/hdtf__cross_identity__id_0027_id_0198.png}
\stackedfigure{cross_identity/hdtf__cross_identity__id_0028_id_0025.png}
\stackedfigure{cross_identity/hdtf__cross_identity__id_0035_id_0155.png}
\stackedfigure{cross_identity/hdtf__cross_identity__id_0036_id_0075.png}
\stackedfigure{cross_identity/hdtf__cross_identity__id_0037_id_0154.png}
\stackedfigure{cross_identity/hdtf__cross_identity__id_0011_id_0051.png}
\stackedfigure{cross_identity/hdtf__cross_identity__id_0042_id_0151.png}
\stackedfigure{cross_identity/hdtf__cross_identity__id_0048_id_0029.png}
\stackedfigure{cross_identity/hdtf__cross_identity__id_0049_id_0183.png}
\stackedfigure{cross_identity/hdtf__cross_identity__id_0065_id_0002.png}
\stackedfigure{cross_identity/hdtf__cross_identity__id_0069_id_0098.png}
\stackedfigure{cross_identity/hdtf__cross_identity__id_0070_id_0055.png}
\stackedfigure{cross_identity/hdtf__cross_identity__id_0074_id_0162.png}
\stackedfigure{cross_identity/hdtf__cross_identity__id_0086_id_0180.png}
\stackedfigure{cross_identity/hdtf__cross_identity__id_0087_id_0132.png}
\stackedfigure{cross_identity/hdtf__cross_identity__id_0099_id_0184.png}
\stackedfigure{cross_identity/hdtf__cross_identity__id_0101_id_0120.png}
\stackedfigure{cross_identity/hdtf__cross_identity__id_0119_id_0087.png}
\stackedfigure{cross_identity/hdtf__cross_identity__id_0121_id_0190.png}
\stackedfigure{cross_identity/hdtf__cross_identity__id_0167_id_0061.png}
\stackedfigure{cross_identity/hdtf__cross_identity__id_0182_id_0034.png}
\stackedfigure{cross_identity/hdtf__cross_identity__id_0190_id_0102.png}

\subsection{Same-Identity Reconstruction}
\label{app:same-identity-examples}
 
\stackedfigure{same_identity/hdtf__same_identity_reconstruction__id_0097.png}
\stackedfigure{same_identity/hdtf__same_identity_reconstruction__id_0111.png}
\stackedfigure{same_identity/hdtf__same_identity_reconstruction__id_0113.png}
\stackedfigure{same_identity/hdtf__same_identity_reconstruction__id_0123.png}
\stackedfigure{same_identity/hdtf__same_identity_reconstruction__id_0146.png}
\stackedfigure{same_identity/hdtf__same_identity_reconstruction__id_0148.png}
\stackedfigure{same_identity/hdtf__same_identity_reconstruction__id_0149.png}
\stackedfigure{same_identity/hdtf__same_identity_reconstruction__id_0155.png}
\stackedfigure{same_identity/hdtf__same_identity_reconstruction__id_0166.png}
\stackedfigure{same_identity/hdtf__same_identity_reconstruction__id_0170.png}
\stackedfigure{same_identity/hdtf__same_identity_reconstruction__id_0197.png}
\stackedfigure{same_identity/hdtf__same_identity_reconstruction__id_0199.png}
\stackedfigure{same_identity/hdtf__same_identity_reconstruction__id_0201.png}

%% =============================================================
%% APPENDIX B — FLAME REPRESENTATION DETAILS
%% =============================================================
\section{FLAME Representation Details}
\label{app:flame-details}
 
\subsection{Fitting Pipeline}
\label{app:flame-fitting}

Loki's FLAME fits are produced by a custom two-phase pipeline. MICA~\cite{mica} is used only to initialise the identity shape, while per-frame expression and pose reconstruction is handled separately.

\paragraph{Phase 1: Dense tracking.}
Each video is first preprocessed: faces are detected with FaceBoxesV2~\cite{faceboxes} and aligned via PIPNet~\cite{pipnet} landmarks; MICA~\cite{mica} estimates an identity shape prior from ArcFace~\cite{arcface} embeddings; and a FaRL~\cite{farl} face-parsing network (with RetinaFace~\cite{retinaface} for face localisation) produces segmentation masks. Two ViT-based transformers (Pixel3DMM~\cite{pixel3dmm}) then predict per-pixel surface normals and per-pixel UV coordinates for each cropped frame, establishing dense 2D-to-3D correspondences against the FLAME template. With these dense maps in hand, a FLAME video tracker optimises identity, expression, jaw, neck, global rotation, translation, and camera intrinsics jointly across batches of 16 frames. The tracking energy combines dense UV correspondence, dense normal-map alignment, silhouette overlap, and sparse landmark terms focused on eye corners, eyelid closure, and iris position to disambiguate gaze and blinks. Strong temporal smoothness priors are applied to all parameters---head rotation and translation in particular are heavily smoothed to suppress jitter. A single global camera is shared across all frames of a clip.

\paragraph{Phase 2: Re-parameterisation.}
The Phase-1 output uses FLAME 2020 with 300 shape and 100 expression coefficients. Because Loki's driver map is defined on FLAME 2023 with 150 shape and 65 expression coefficients (the FlowFace/CAP4D~\cite{cap4d} convention, in which the jaw is absorbed into the expression basis), a second-stage re-fit converts between parameterisations. This stage minimises a per-vertex weighted L2 distance between the Phase-1 posed meshes and the re-parameterised FLAME, optimising one global identity vector plus per-frame expression, head rotation, translation, eye rotation, and neck rotation. A 500-step pose warmup freezes expression, shape, and eye rotation so that head pose locks in first; the remaining iterations then release all parameters. After convergence, the optimised eye rotations are discarded and replaced with predictions from an L2CS-Net~\cite{l2csnet} gaze model, converted into eyeball rotations in the head's local frame and temporally smoothed with a Gaussian filter ($\sigma = 2$ frames). Background segmentation is produced separately by RobustVideoMatting~\cite{robustvideomatting}.

\paragraph{Output.}
The final per-clip bundle contains: per-frame head rotation, translation, expression ($\vec{\psi} \in \mathbb{R}^{65}$), neck rotation, and gaze-corrected eye rotation; a shared identity shape ($\vec{\beta} \in \mathbb{R}^{150}$); rescaled camera intrinsics and extrinsic in OpenCV convention; and per-frame foreground alpha mattes.
 
\paragraph{Inner-mouth extension.}
The standard FLAME mesh does not model the oral cavity---when the jaw opens, the interior of the mouth is absent. We extend the mesh with 200 procedural inner-mouth vertices arranged as a half-sphere, positioned from the jaw joint and lip vertex 3533, following the approach of CAP4D~\cite{cap4d}. These vertices are \emph{not} part of FLAME's linear blend skinning and do not carry expression deformation (their deformation channels are zero); they contribute only positional encoding channels, providing a static cavity surface inside the moving lips.

\paragraph{Rasterisation.}
The FLAME mesh (5{,}023 base vertices + 200 inner-mouth vertices = 5{,}223 total) is rasterised at $512 \times 512$ using a single PyTorch3D hard-rasterisation call. The live per-frame NDC vertices (computed from FLAME with the current $\vec{\beta}$, $\vec{\psi}$, $\vec{\theta}$, and camera) determine rasterisation geometry only---which triangle face is visible at each pixel and the barycentric interpolation weights. The \emph{values} written to the output channels are the template-space positional encodings and expression deformations, not the live vertex positions. This separation is what ensures the positional encoding channels are identity-invariant: the live vertices control \emph{where} the values appear in the image, but the values themselves are fixed properties of the canonical mesh.

\subsection{Driver Map Channel Specification}
\label{app:driver-map-channels}
 
The driver map has shape $(B, T, 45, 512, 512)$ and decomposes as follows:
 
\begin{center}
\small
\begin{tabular}{clcl}
\toprule
Channels & Content & Depends on & Invariant to \\
\midrule
1--42 & Sinusoidal positional encoding $\gamma(\mathbf{p})$ & Template positions & Identity, expression \\
43--45 & Expression deformation $(\Delta x, \Delta y, \Delta z)$ & $\vec{\psi}$ only & Identity, pose \\
\midrule
--- & Off-mesh pixels & \multicolumn{2}{c}{Zero in all 45 channels} \\
\bottomrule
\end{tabular}
\end{center}
 
\begin{figure}[h]
\centering
\includegraphics[width=0.85\textwidth]{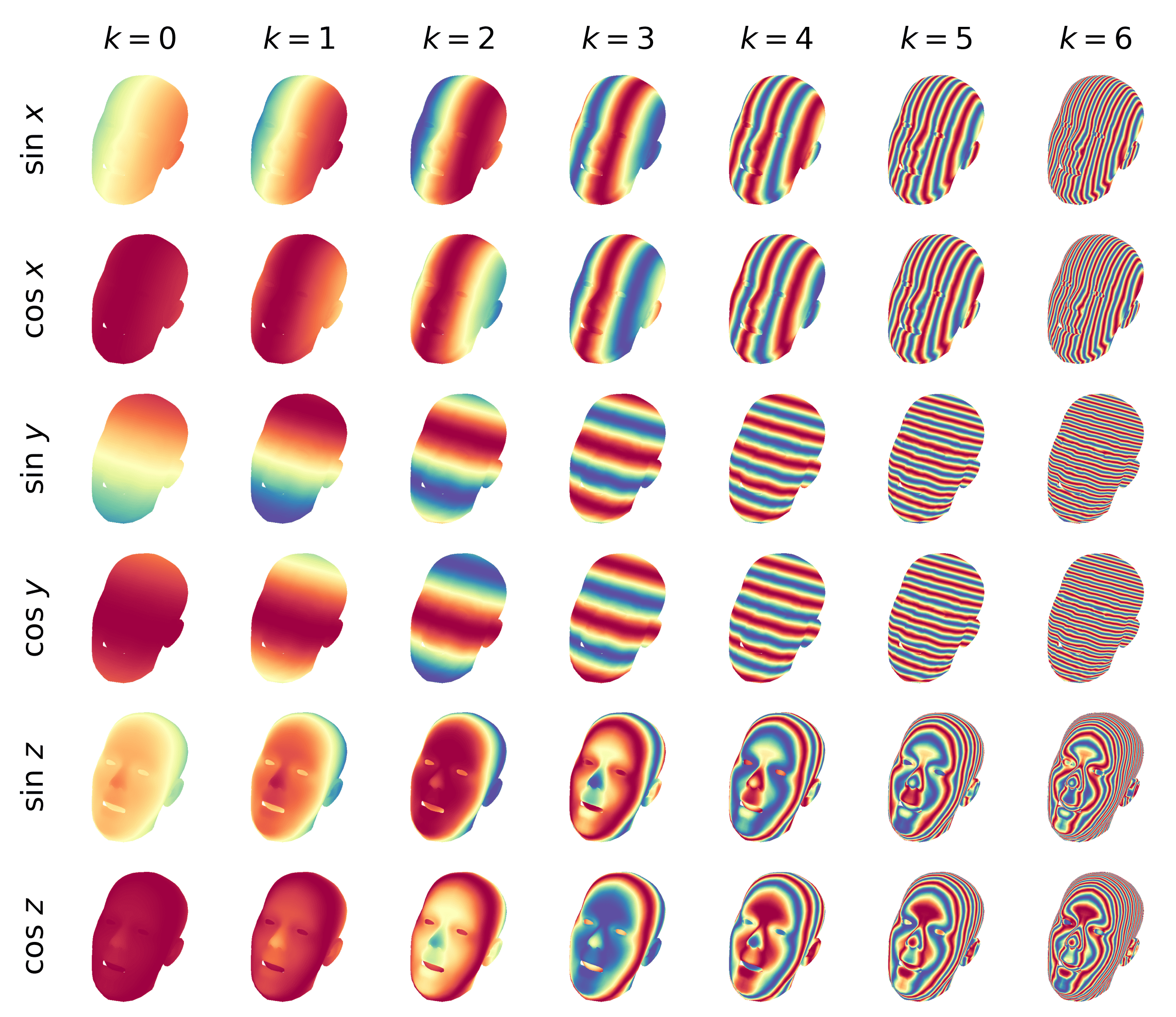}
\caption{\textbf{Full 42-channel positional-encoding visualisation}. Rows correspond to the six axis-grouped channel blocks ($\sin x$, $\cos x$, $\sin y$, $\cos y$, $\sin z$, $\cos z$); columns to the seven octave frequencies ($k = 0, \ldots, 6$). Low-$k$ channels produce smooth gradients that distinguish broad face regions; high-$k$ channels produce fine fringes that give each point a unique signature.}
\label{fig:app-posenc-full}
\end{figure}

\paragraph{Positional encoding channels (1--42).}
The 42 channels encode FLAME's template-space vertex positions through sinusoidal positional encoding at 7 octave-spaced frequencies:
\begin{equation}
\gamma(\mathbf{p}) = \bigl[\sin(2^0 \mathbf{p}),\; \cos(2^0 \mathbf{p}),\; \ldots,\; \sin(2^6 \mathbf{p}),\; \cos(2^6 \mathbf{p})\bigr] \in \mathbb{R}^{42},
\end{equation}

where $\mathbf{p} = (x, y, z)$ are the template vertex coordinates, mean-centred and max-normalised once at initialisation. The channels are organised by spatial axis: all sine components for $x$ across the 7 frequencies appear first, followed by all cosine components for $x$, then the corresponding groups for $y$, and finally those for $z$. This axis-grouped ordering yields 6 blocks of 7 channels each, totalling 42 channels as shown in each panel in Figure~\ref{fig:app-posenc-full}. Rows correspond to the six axis-grouped channel blocks ($\sin x$, $\cos x$, $\sin y$, $\cos y$, $\sin z$, $\cos z$); columns correspond to the seven octave frequencies ($k = 0, \ldots, 6$), giving the full 6 × 7 = 42 channels supplied as positional features to the conditioning encoder. Reading left to right, low-frequency channels ($k = 0$) produce smooth gradients that distinguish broad face regions — for example, $\sin x$ at $k = 0$ separates the left and right halves of the head — while high-frequency channels ($k = 6$) produce dense fringes that give each surface point a unique signature. Reading top to bottom, the three coordinate axes visibly encode different geometric structure: $x$-rows trace horizontal bands, $y$-rows trace vertical bands, and $z$-rows highlight depth-driven features such as the protruding nose, brow ridges, and cheekbones. Crucially, the values written into each pixel are computed from the canonical FLAME template coordinates rather than the live per-frame vertex positions, so the same anatomical point — the nose tip, the corner of the lip — always produces the same value across identities, expressions, and head poses; only the spatial layout in image space changes with pose. This identity-invariance is what allows the encoder to learn a stable correspondence between FLAME geometry and image-space features.

\paragraph{Expression deformation channels (43--45).}
The three deformation channels carry the per-vertex expression offset $\Delta_{\text{expr}} = B_E(\vec{\psi};\, \mathbf{E})$, barycentrically interpolated across triangle faces during rasterisation. The offsets are normalised by a fixed scalar of $0.0104$ (the standard deviation of expression deformation across the training corpus) before rasterisation, so their magnitude is compatible with the positional encoding channels which lie in $[-1, 1]$.
 
\paragraph{Mask.}
Only head and inner-mouth vertices contribute to the rasterisation. All off-mesh pixels (background) are zero across all 45 channels.

%% =============================================================
%% APPENDIX C — GENERATION BACKBONE DETAILS
%% =============================================================
\section{Generation Backbone Details}
\label{app:backbone-details}
 
\subsection{Noise Schedule}
\label{app:noise-schedule}
 
Loki uses a linear noise schedule in which the noise magnitude increases steadily across diffusion steps, with an enforce-zero-terminal-SNR correction~\cite{enforcezerosnr} that ensures the final diffusion step reaches pure noise. Given a clean latent $\mathbf{z}_0$, a noise sample $\varepsilon \sim \mathcal{N}(\mathbf{0}, \mathbf{I})$, and a timestep $t$, the noisy latent is
\begin{equation}
\mathbf{z}_t = \sqrt{\bar{\alpha}_t}\,\mathbf{z}_0 + \sqrt{1 - \bar{\alpha}_t}\,\varepsilon.
\end{equation}
 
The training objective is $\varepsilon$-prediction MSE over the $T$ target latent slots:
\begin{equation}
\mathcal{L} = \mathbb{E}_{\mathbf{z}_0,\,\varepsilon,\,t}\bigl[\,\|\varepsilon - \varepsilon_\phi(\mathbf{z}_t,\, t,\, \mathbf{c})\|^2\,\bigr],
\end{equation}
where $\mathbf{c}$ denotes the full conditioning (driver map and reference features) and $\varepsilon_\phi$ is the generation UNet. Per-frame timesteps are drawn independently from $[0, 1000)$ on the joint $(B, T)$ grid.
 
\paragraph{Temporal shift.}
Following~\cite{hoogeboom2023simple}, who show that the noise schedule must be shifted when signal dimensionality increases, we apply an analogous shift along the temporal axis: jointly denoising $T$ frames increases the effective signal, requiring a shift with ratio $\sqrt{1/n_{\text{gen}}}$ where $n_{\text{gen}} = T - 1 = 15$, giving a shift ratio of approximately $0.258$.
 
\paragraph{Inference.}
At inference, clean latents are recovered by DDIM sampling~\cite{ddim} with $50$ steps and $\eta = 0$ (deterministic):
\begin{equation}
\mathbf{z}_{t-1}
= \sqrt{\bar{\alpha}_{t-1}}\,\hat{\mathbf{z}}_0
+ \sqrt{1 - \bar{\alpha}_{t-1}}\,\varepsilon_\phi(\mathbf{z}_t,\, t,\, \mathbf{c}),
\quad
\hat{\mathbf{z}}_0
= \frac{\mathbf{z}_t - \sqrt{1 - \bar{\alpha}_t}\,\varepsilon_\phi(\mathbf{z}_t,\, t,\, \mathbf{c})}{\sqrt{\bar{\alpha}_t}},
\end{equation}
with classifier-free guidance~\cite{cfg} at scale $s = 2.0$:
\begin{equation}
\tilde{\varepsilon} = \varepsilon_{\text{uncond}} + s\,(\varepsilon_{\text{cond}} - \varepsilon_{\text{uncond}}).
\end{equation}

\subsection{Attention Token Layouts}
\label{app:attention-shapes}
 
The generation UNet processes $T = 16$ frames jointly. The attention mode and resulting sequence lengths vary by resolution stage.
 
\paragraph{Per-frame spatial attention ($64 \times 64$).}
At the outermost resolution, each frame is attended independently. For a single frame, the spatial sequence has $64 \times 64 = 4{,}096$ tokens. Reference features from the identity path are appended as additional key-value tokens, giving a key-value sequence of $4{,}096 + 4{,}096 = 8{,}192$ tokens per frame. The query sequence remains $4{,}096$. This mode is applied independently to each of the $T$ frames.
 
\paragraph{3D spatiotemporal attention ($32 \times 32$, $16 \times 16$, $8 \times 8$).}
At the three inner resolutions, all frames are concatenated into a single spatiotemporal sequence before attention:
 
\begin{center}
\small
\begin{tabular}{lcccc}
\toprule
Resolution & Spatial tokens & Temporal tokens & Query length & KV length (with ref) \\
\midrule
$32 \times 32$ & 1{,}024 & $16 \times 1{,}024 = 16{,}384$ & 16{,}384 & $16{,}384 + 1{,}024$ \\
$16 \times 16$ & 256 & $16 \times 256 = 4{,}096$ & 4{,}096 & $4{,}096 + 256$ \\
$8 \times 8$ & 64 & $16 \times 64 = 1{,}024$ & 1{,}024 & $1{,}024 + 64$ \\
\bottomrule
\end{tabular}
\end{center}
 
Reference key-value tokens are appended once per batch element (not repeated across frames), so every query token---at every spatial position and every frame---can attend to the reference's identity features.
 
\paragraph{Reshape operations.}
For 3D spatiotemporal attention, the input tensor of shape $(B, T, C, H, W)$ is reshaped to $(B, T \cdot H \cdot W, C)$ before the attention call. For per-frame spatial attention, the tensor is reshaped to $(B \cdot T, H \cdot W, C)$, so each frame is processed independently. After attention, the inverse reshape restores the original layout. All attention computations use \texttt{xformers} memory-efficient attention~\cite{xformers}.

\subsection{Conditioning Encoder Architecture}
\label{app:cond-encoder}
 
The conditioning encoder bridges the $45$-channel driver map at $512 \times 512$ to the generation UNet's $64 \times 64 \times 320$ feature space. Its design mirrors the convolutional style of Stable Diffusion 2.1 to ensure statistical compatibility between the encoder's output features and the UNet's input expectations.
 
\paragraph{Architecture.}
\begin{center}
\small
\begin{tabular}{llcc}
\toprule
Stage & Operation & Output resolution & Output channels \\
\midrule
Stem & $3 \times 3$ Conv, SiLU & $512 \times 512$ & 64 \\
Down 1 & ResBlock + stride-2 Conv & $256 \times 256$ & 128 \\
Down 2 & ResBlock + stride-2 Conv & $128 \times 128$ & 256 \\
Down 3 & ResBlock + stride-2 Conv & $64 \times 64$ & 320 \\
Refine & ResBlock & $64 \times 64$ & 320 \\
Output & GroupNorm $\to$ SiLU $\to$ $3 \times 3$ Conv (zero-init) & $64 \times 64$ & 320 \\
\bottomrule
\end{tabular}
\end{center}
 
Each ResBlock follows the SD 2.1 pattern: GroupNorm $\to$ SiLU $\to$ $3 \times 3$ Conv $\to$ GroupNorm $\to$ SiLU $\to$ $3 \times 3$ Conv, with a residual skip connection (and a $1 \times 1$ projection when channel dimensions change). Each downsampling stage uses a single stride-2 convolution to reduce the spatial resolution by a factor of two.

\paragraph{Zero initialisation.}
The final $3 \times 3$ convolution has both its weight and bias initialised to zero. At the start of training, the encoder therefore contributes nothing to the generation UNet's feature map, which begins from its pretrained Stable Diffusion state. The conditioning signal is learned gradually as training progresses---the same stability mechanism used by ControlNet~\cite{controlnet}.
 
\paragraph{Parameter count.}
The conditioning encoder contains approximately 7.35M trainable parameters, compared to ${\sim}822.9$M in the generation UNet.

\subsection{Identity Path Implementation}
\label{app:identity-path}
 
\paragraph{Reference UNet.}
The reference UNet is a frozen copy of the Stable Diffusion 2.1 UNet, initialised from the same pretrained checkpoint as the generation UNet. It retains cross-attention layers architecturally, but these attend to the OpenCLIP~\cite{openclip} encoding of the empty string, computed once at initialisation. No textual prompt conditions either the reference or the generation process.
 
\paragraph{Feature extraction.}
The VAE-encoded reference image ($64 \times 64 \times 4$) is passed through the reference UNet once per sample. At each self-attention layer, a forward hook captures the post-LayerNorm activation tensor before the attention computation. These captured tensors serve as the identity features that are injected into the generation UNet.
 
\paragraph{Key-value injection.}
At each matching self-attention layer in the generation UNet, the captured reference features are concatenated with the generation features along the token dimension before key and value projection:
\begin{equation}
K = W_K [\mathbf{x}_{\text{gen}};\, \mathbf{x}_{\text{ref}}], \quad V = W_V [\mathbf{x}_{\text{gen}};\, \mathbf{x}_{\text{ref}}],
\end{equation}
where $W_K$ and $W_V$ are the generation UNet's own projection matrices---no separate learned projections are introduced. The query is computed from generation features only: $Q = W_Q \mathbf{x}_{\text{gen}}$.

\begin{figure}[h]
\centering
\includegraphics[width=\textwidth]{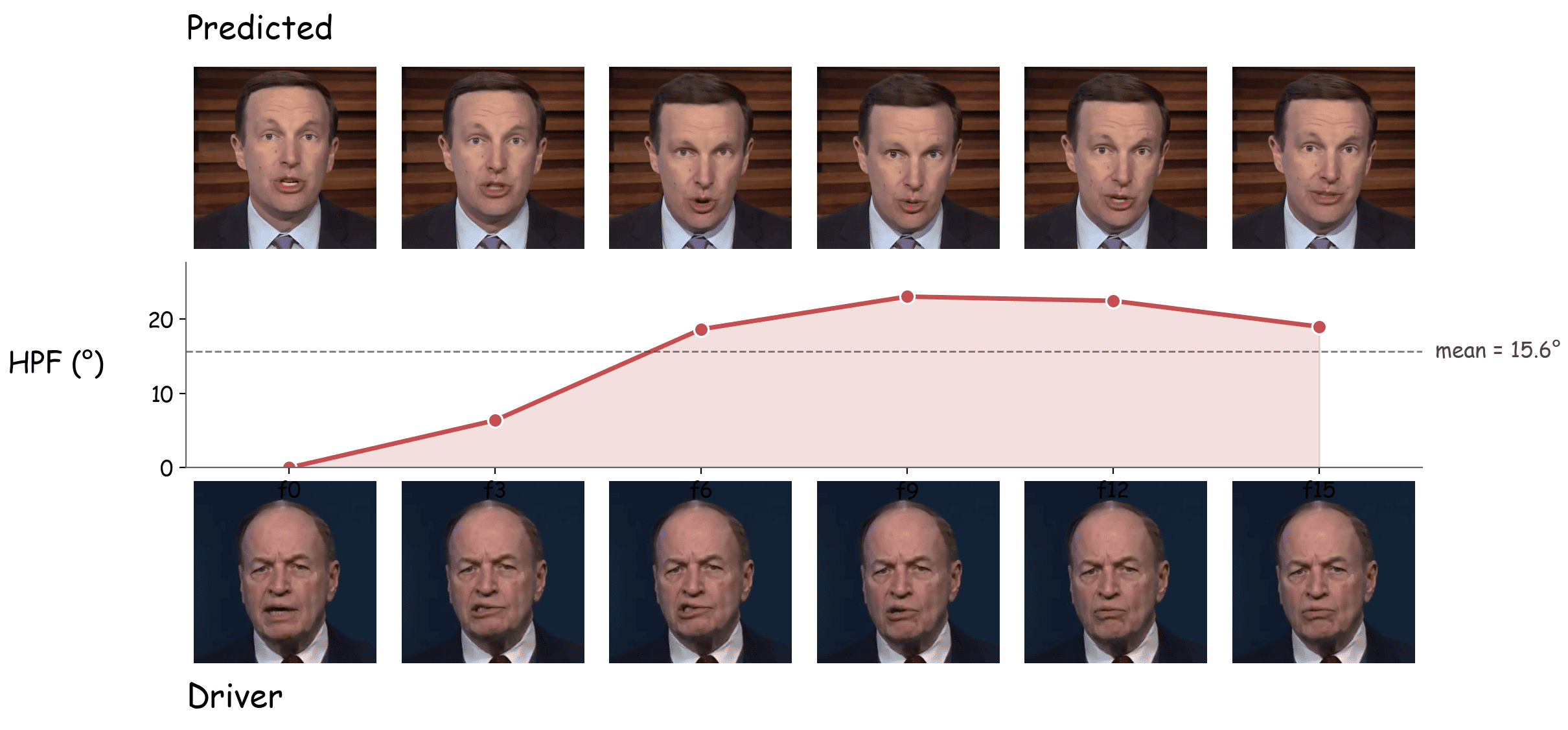}
\caption{\textbf{Per-frame head-pose follow on a failure case generated by EchoMimic}. Six evenly-sampled frames from the 16-frame inference window of EchoMimic on cross ID sample from HDTF~\cite{hdtf}. The top row shows EchoMimic's~\cite{echomimic} predicted output, the bottom row shows the driver clip (the motion source the prediction is supposed to follow), and the middle panel plots the per-frame Head-Pose-Follow (HPF) error in degrees---the geodesic angular distance between predicted and driver head rotation at frame $t$, anchored at frame 0. Mean HPF across the window is $15.6^\circ$.}
\label{fig:hpf-echomimic-failure}
\end{figure}

\paragraph{Handling temporal modes.}
In 3D spatiotemporal attention layers (resolutions $32^2$, $16^2$, $8^2$), the reference features are a single-frame spatial tensor. They are appended once to the key-value sequence; every query token across all $T$ frames attends to the same reference features. In per-frame spatial attention layers ($64^2$ resolution), the reference features are repeated $T$ times so that each frame's independent attention call includes the reference.
 
\paragraph{Caching.}
The reference UNet forward pass is executed once per sample at the start of inference. The extracted per-layer features are cached and reused at every denoising step, contributing no additional compute beyond the initial extraction. During training, the reference UNet forward pass is included in the computation graph but receives no gradients (all parameters frozen).

%% =============================================================
%% APPENDIX D — EVALUATION METRIC DERIVATIONS
%% =============================================================

\section{Evaluation Metric Derivations}
\label{app:metric-derivations}

\subsection{Head Pose Follow (HPF)}
\label{app:hpf}

HPF measures how faithfully the generated clip reproduces the driver's head-rotation trajectory, using FLAME's own pose parameters as the representation space.

\paragraph{Head rotation from FLAME.}
FLAME exposes head pose through two rotation parameters: a global rigid rotation $\vec{r}_{\text{global}} \in \mathbb{R}^3$ applied to the entire mesh after linear-blend skinning, and a neck-joint rotation $\vec{r}_{\text{neck}} \in \mathbb{R}^3$ applied within LBS at the skeleton's root joint. These two quantities are formally redundant in FLAME's head-only model---the same visible head orientation can be reached by many $(\vec{r}_{\text{global}}, \vec{r}_{\text{neck}})$ allocations---and a viewer of the rendered video sees only their combined effect. HPF therefore composes them into a single visible head-rotation matrix per frame:
\begin{equation}
R_{\text{head}}[t] = R(\vec{r}_{\text{global}}[t]) \cdot R(\vec{r}_{\text{neck}}[t]),
\end{equation}
where $R(\cdot)$ converts an axis-angle vector to a $3 \times 3$ rotation matrix via Rodrigues' formula. This composition reflects FLAME's rendering order---the neck rotation is applied first within LBS, then the global rotation rotates the whole mesh---so $R_{\text{head}}[t]$ encodes the orientation of any head-attached point (nose tip, eye corner, ear) in the FLAME reconstruction's per-clip camera frame.

\paragraph{Frame-0-anchored delta rotations.}
Because the predicted and target clips are reconstructed independently, their absolute head orientations live in unrelated camera frames. Comparing $R_{\text{head}}^{\text{pred}}[t]$ to $R_{\text{head}}^{\text{tgt}}[t]$ directly would conflate per-clip camera offset with genuine pose error. We cancel that offset by computing frame-0-anchored deltas:
\begin{equation}
\Delta R[t] = R_{\text{head}}[t] \cdot R_{\text{head}}[0]^\top,
\end{equation}
which is the rotation that takes the head from its $t{=}0$ orientation to its time-$t$ orientation. By construction $\Delta R[0] = I$ for both prediction and target, and any constant camera offset drops out symmetrically. The result is a pure pose trajectory: the sequence of rotations the head induces relative to its starting position.

We use cumulative anchoring to frame 0 rather than inter-frame deltas ($R_{\text{head}}[t] \cdot R_{\text{head}}[t{-}1]^\top$) for three reasons. First, cumulative deltas penalise trajectory divergence---if the prediction drifts steadily off the driver's path, error grows over time. Inter-frame deltas would let slow systematic drift go unpenalised: a prediction adding $0.3°$ of yaw every frame would match the driver's inter-frame rotation at each step yet end the clip $4.8°$ off target. Second, cumulative deltas have better signal-to-noise ratio: at 25\,fps the head moves ${\sim}1$--$2°$ per frame while the FLAME reconstruction carries ${\sim}0.5°$ per-frame noise, giving inter-frame SNR of ${\sim}2$--$4\times$; cumulative deltas reach $10$--$20°$ over the 16-frame window, raising SNR to $20$--$40\times$. Third, frame 0 is a well-anchored reference in both evaluation protocols: for same-identity reconstruction, the predicted and ground-truth frame 0 correspond by construction; for cross ID reenactment, both are generated from the same driver frame 0.

\paragraph{Geodesic angular distance.}
At each frame $t$, the predicted and target delta rotations are converted to unit quaternions and compared via the geodesic distance on $\mathrm{SO}(3)$:
\begin{equation}
\theta[t] = 2 \cdot \arccos\!\bigl(\text{clip}\bigl(|\mathbf{q}_{\text{pred}}[t] \cdot \mathbf{q}_{\text{tgt}}[t]|,\; -1,\; 1\bigr)\bigr),
\end{equation}
reported in degrees. This is the smallest rotation angle that would bring one head's orientation to the other's---a single, coordinate-frame-invariant scalar. The absolute value handles the quaternion double cover ($\mathbf{q}$ and $-\mathbf{q}$ represent the same rotation), and the clipping prevents numerical issues at the boundaries of $\arccos$.

We report the geodesic distance as a single scalar rather than a per-axis yaw/pitch/roll breakdown. Euler-angle comparisons are unreliable: different decomposition conventions produce different numbers, wrapping at $\pm 180°$ introduces discontinuities, and the same rotation has multiple equivalent Euler triples near gimbal lock. The geodesic distance avoids all of these pathologies.

\paragraph{Aggregation.}
The per-sample HPF is the arithmetic mean of $\theta[t]$ over the $T = 16$-frame evaluation window, with $\theta[0] = 0$ contributing trivially by construction. The dataset-level HPF is the mean across all evaluation samples. The result is a single degree-valued, lower-is-better number: single-digit values indicate tight pose-trajectory following, while double-digit values indicate the head is going somewhere measurably different from the driver's trajectory.

\subsection{Expression Follow (HEF)}
\label{app:hef}
 
\paragraph{Motivation.}
HEF measures whether the generated facial expression matches the driver's---lip shape, jaw drop, brow raise, eye gaze---while deliberately ignoring head pose, identity shape, and camera. Direct comparison of expression coefficient vectors $\vec{\psi}$ across independently reconstructed clips is unreliable because the visible effect of an expression coefficient depends on the head's pose and the camera's viewpoint. Two identical $\vec{\psi}$ vectors can produce visually different expressions when rendered from different angles.

\begin{figure}[h]
\centering
\includegraphics[width=\textwidth]{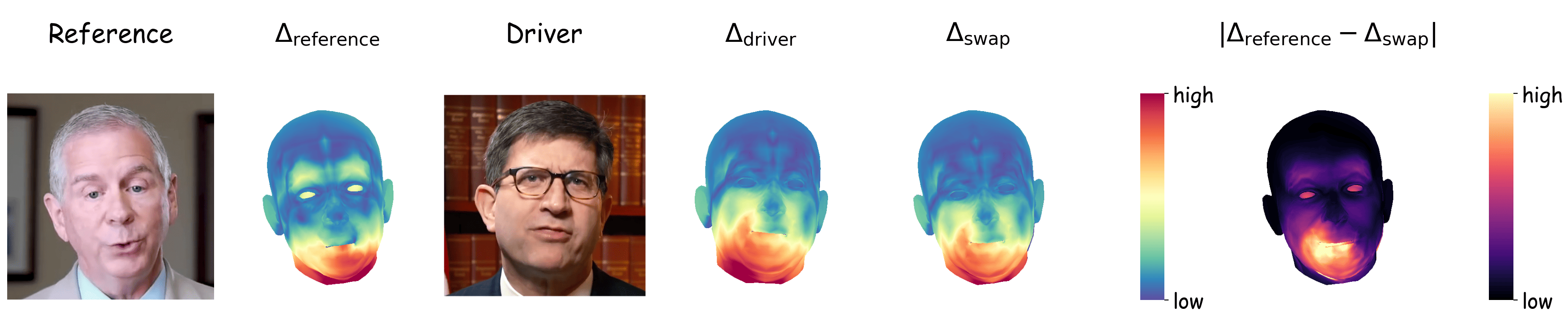}
\caption{\textbf{Visualization of HEF error on a single retargeted pair}. Left to right: the reference frame and its rasterized expression-deformation map ($\Delta_{\mathrm{reference}}$); the driver frame and its own expression map ($\Delta_{\mathrm{driver}}$); the swapped map ($\Delta_{\mathrm{swap}}$), produced by inserting the driver's expression coefficients into the reference's pose, shape, and camera; and the per-pixel L1 difference $|\Delta_{\mathrm{reference}} - \Delta_{\mathrm{swap}}|$ that the metric integrates over on-mesh pixels. All three $\Delta$ maps share a single magnitude scale so they are directly comparable; the error map is given on the rightmost column. Because pose and shape are held to the reference's, the residual between $\Delta_{\mathrm{reference}}$ and $\Delta_{\mathrm{swap}}$ is attributable to expression alone — concentrated here in the lower face, where the driver's mouth and jaw configuration disagree with the reference's.}
\label{fig:hef_l1_visualization}
\end{figure}

\paragraph{Expression substitution.}
We resolve this by comparing expression effects in a shared rendering context. At each frame $t$:
\begin{enumerate}
    \item Take the target's full FLAME fit verbatim: identity $\vec{\beta}_{\text{tgt}}$, pose $\vec{\theta}_{\text{tgt}}$, camera $\text{cam}_{\text{tgt}}$.
    \item Render two deformation maps from this shared context:
    \begin{itemize}
        \item \textbf{Target map}: using the target's expression parameters ($\vec{\psi}_{\text{tgt}}$, $\text{eye\_rot}_{\text{tgt}}$, $\text{jaw\_rot}_{\text{tgt}}$).
        \item \textbf{Substituted map}: replacing only the expression parameters with the prediction's ($\vec{\psi}_{\text{pred}}$, $\text{eye\_rot}_{\text{pred}}$, $\text{jaw\_rot}_{\text{pred}}$).
    \end{itemize}
\end{enumerate}
Because everything except expression is held identical (same $\vec{\beta}$, same $\vec{\theta}$, same camera), both maps occupy exactly the same pixels in image space. Any difference between them is attributable purely to expression mismatch.

\paragraph{Deformation maps.}
Each map contains the three expression deformation channels $(\Delta x, \Delta y, \Delta z)$ rasterised through the FLAME mesh, identical in format to the deformation channels of the driver map described in \S\ref{sec:method-driver}. The rasterisation uses the target's identity shape and pose for both maps---only the expression deformation offsets differ.

\paragraph{Mask-aware L1.}
The per-frame score is the mean absolute difference across the three deformation channels, averaged over on-mesh pixels only:
\begin{equation}
\text{HEF}[t] = \frac{1}{|\mathcal{M}|} \sum_{(i,j) \in \mathcal{M}} \frac{1}{3} \sum_{c \in \{x,y,z\}} \bigl| \Delta_c^{\text{tgt}}[i,j] - \Delta_c^{\text{sub}}[i,j] \bigr|,
\end{equation}
where $\mathcal{M}$ is the set of on-mesh pixels (non-zero in both maps by construction). Background pixels are excluded to prevent dilution of the score.
 
\paragraph{Why L1, not L2.}
L1 treats all expression deviations equally regardless of magnitude, while L2 disproportionately penalises large deviations. In portrait animation, we care about the full spectrum of expression mismatch---a system that consistently produces slight jaw under-opening across all frames is as problematic as one that occasionally misses a large expression---and L1 captures this more evenly.
 
\paragraph{Aggregation.}
The per-sample HEF is the mean of $\text{HEF}[t]$ over the $T = 16$-frame evaluation window. The dataset-level HEF is the mean across all evaluation clips.

\subsection{HEF Calibration Study}
\label{app:hef-calibration}

To give HEF scores concrete meaning, we establish four empirical reference points spanning the metric's range. All measurements use our subset of TalkVid~\cite{talkvid} corpus (10{,}650 FLAME-fitted clips), which provides sufficient sample size to stabilise the distributional tails. Results on HDTF agree to within sampling noise.

Each reference point is measured at single-frame granularity rather than after the 16-frame temporal averaging used at evaluation time. This isolates per-frame expression fidelity and ensures the reported anchors are not deflated by temporal smoothing.

\paragraph{Theoretical lower bound (0.000).}
Pairing any frame's FLAME fit against itself produces HEF $= 0.000$ exactly, confirming that the rasterisation pipeline introduces no spurious residuals.

\paragraph{Practical lower bound (${\sim}$0.03).}
Two frames within a few-frame window of the same clip have nearly identical expression---at 25\,fps, natural articulation changes the expression vector negligibly over 40\,ms. Because the HEF substitution holds identity, pose, and camera fixed, any residual reflects imprecision in the FLAME reconstruction rather than genuine expression change. This is the expected score for a model that perfectly reproduces the target's expression.

\paragraph{No-skill baseline ($0.103 \pm 0.040$).}
Fifty pairs are drawn by selecting two distinct clips uniformly at random and sampling one frame from each. The mean HEF across these pairs is $0.103$ with a standard deviation of $0.040$ (standard error ${\approx}0.006$). Since the substitution mechanism removes the effect of identity, pose, and camera, this number reflects the expected distance between two expression vectors drawn independently from the empirical distribution. A system scoring at or above $0.10$ has not demonstrated expression transfer above chance.

\paragraph{Realistic upper bound (${\sim}$0.18).}
The most neutral frame in the corpus (lowest $\|\vec{\psi}\|_2$ across all clips and frames) is paired against the most expressive frame at the 99th percentile of per-clip maximum $\|\vec{\psi}\|_2$, yielding HEF $= 0.181$. We use the 99th percentile rather than the global maximum because the tail of the expression-norm distribution is dominated by FLAME reconstruction failures---the global maximum $\|\vec{\psi}\|_2$ exceeds the p99 value by roughly $6\times$ and produces an HEF of ${\sim}0.67$, far outside the plausible range for valid fits. The p99 cutoff anchors the upper bound to the most expressive frame a model is likely to encounter in practice, rather than to a reconstruction artefact.

\paragraph{Interpretation.}
The four anchors partition HEF into interpretable regions:

\begin{center}
\small
\begin{tabular}{ll}
\toprule
HEF range & Interpretation \\
\midrule
$0.00$--${\sim}0.03$ & Reconstruction noise floor \\
${\sim}0.03$--${\sim}0.10$ & Meaningful expression transfer \\
${\sim}0.10$--${\sim}0.18$ & Between no-skill baseline and realistic ceiling \\
$> {\sim}0.18$ & Typically indicates reconstruction failure \\
\bottomrule
\end{tabular}
\end{center}

The plausible range is narrow (${\sim}0.00$ to ${\sim}0.20$) because HEF operates in normalised deformation space---expression offsets divided by $0.0104$ before rasterisation---and most of the face deforms little, with expression variance concentrated around the mouth and eyes. Within this range, relative gaps are highly informative: Loki's cross ID HEF of $0.083$ falls in the meaningful-transfer region, while the audio-driven baselines at $0.11$--$0.12$ are at or above the no-skill boundary.

\bibliographystyle{unsrtnat}

\bibliography{references}

\newpage
\section*{NeurIPS Paper Checklist}

\begin{enumerate}

\item {\bf Claims}
    \item[] Question: Do the main claims made in the abstract and introduction accurately reflect the paper's contributions and scope?
    \item[] Answer: \answerYes{}
    \item[] Justification: The abstract and introduction state three contributions: (1) a portrait animation framework with disentangled parametric conditioning reducing trainable parameters by ${\sim}3.5\times$, (2) a training recipe enabling cross-identity inference from same-identity training via parametric substitution, and (3) two evaluation metrics for head pose and expression fidelity. All three are supported by the experimental results in \S\ref{sec:exp-results}: Tables~\ref{tab:results} and~\ref{tab:ablation} verify the quantitative claims, and the cross-identity capability is demonstrated qualitatively in Figures~\ref{fig:teaser} and~\ref{fig:qualitative}. Limitations including dependence on FLAME fitting quality and the pixel-metric gap are discussed in the conclusion.

\item {\bf Limitations}
    \item[] Question: Does the paper discuss the limitations of the work performed by the authors?
    \item[] Answer: \answerYes{}
    \item[] Justification: The conclusion explicitly identifies two limitations: (1) Loki's conditioning is bounded by the fidelity of the underlying FLAME fits---expressions outside the model's basis, occluded faces, and fitting failures propagate directly into the driver map; and (2) the pixel-metric gap relative to RGB-conditioned baselines remains open. The results section further discusses how FLAME camera estimation (independent of ground-truth framing) accounts for the gap on pixel-aligned metrics.

\item {\bf Theory assumptions and proofs}
    \item[] Question: For each theoretical result, does the paper provide the full set of assumptions and a complete (and correct) proof?
    \item[] Answer: \answerYes{}
    \item[] Justification: The paper's theoretical claim is that FLAME's parameter independence (Eq.~\ref{eq:flame-tp}) guarantees that cross-identity reenactment produces a driver map identical in content to same-identity training. The assumption---that FLAME's identity, expression, and pose offset fields are independent linear combinations over orthonormal bases---is stated explicitly in \S\ref{sec:method-driver}. The derivation in the ``From same-identity training to cross-identity inference'' paragraph walks through the substitution step by step, showing that positional encoding and expression deformation channels are unchanged under identity substitution, with only the spatial layout affected by the rasterisation geometry.

    \item {\bf Experimental result reproducibility}
    \item[] Question: Does the paper fully disclose all the information needed to reproduce the main experimental results of the paper to the extent that it affects the main claims and/or conclusions of the paper (regardless of whether the code and data are provided or not)?
    \item[] Answer: \answerYes{}
    \item[] Justification: The paper specifies the backbone architecture (Stable Diffusion 2.1), the FLAME parameterisation and rasterisation procedure (\S\ref{sec:method-driver}), the identity path mechanism (\S\ref{sec:method-identity}), the conditioning encoder design (\S\ref{sec:method-backbone}), training hyperparameters (batch size, learning rate, optimizer, number of steps, GPU configuration in \S\ref{sec:exp-training}), inference procedure (DDIM steps, guidance scale), dataset details (TalkVid subset size, HDTF evaluation protocol), and metric definitions (HPF, HEF in \S\ref{sec:exp-metrics}). Additional architectural details are provided in the appendix. All baselines use official code and released checkpoints.

\item {\bf Open access to data and code}
    \item[] Question: Does the paper provide open access to the data and code, with sufficient instructions to faithfully reproduce the main experimental results, as described in supplemental material?
    \item[] Answer: \answerNo{}
    \item[] Justification: Code and model checkpoints are not released at submission time to preserve anonymity. The paper provides sufficient architectural and training detail for independent reimplementation. We intend to release code and pretrained weights upon acceptance.

\item {\bf Experimental setting/details}
    \item[] Question: Does the paper specify all the training and test details (e.g., data splits, hyperparameters, how they were chosen, type of optimizer) necessary to understand the results?
    \item[] Answer: \answerYes{}
    \item[] Justification: \S\ref{sec:exp-setup} specifies the dataset splits (1{,}493 training / 83 validation identities from TalkVid, 212 held-out identities from HDTF) and evaluation protocols (same-identity and cross-identity). \S\ref{sec:exp-training} specifies the optimizer (AdamW), learning rate ($10^{-4}$), batch size (4 per GPU), number of GPUs ($8\times$H200), training duration (30k steps), clip length ($T{=}16$), classifier-free guidance dropout ($p{=}0.1$), and inference settings (DDIM 50 steps, guidance scale $s{=}2.0$). The validation set is used for hyperparameter tuning only. Additional details are provided in the appendix.

\item {\bf Experiment statistical significance}
    \item[] Question: Does the paper report error bars suitably and correctly defined or other appropriate information about the statistical significance of the experiments?
    \item[] Answer: \answerNo{}
    \item[] Justification: Results are reported as single runs without error bars. Training a diffusion model for 30k steps on $8\times$H200 GPUs is computationally expensive, making multiple independent runs impractical. However, all metrics are averaged over all 212 clips in the HDTF evaluation set, providing a reasonable estimate of expected performance. The HEF calibration study (Appendix~\ref{app:hef-calibration}) provides reference points (noise floor, random baseline, upper bound) to contextualise the reported values.

\item {\bf Experiments compute resources}
    \item[] Question: For each experiment, does the paper provide sufficient information on the computer resources (type of compute workers, memory, time of execution) needed to reproduce the experiments?
    \item[] Answer: \answerYes{}
    \item[] Justification: \S\ref{sec:exp-training} reports the GPU type ($8\times$H200), batch size (4 per GPU), and training duration (30k steps). The trainable parameter counts are stated for both the generation UNet (${\sim}823$M) and the conditioning encoder (${\sim}7.35$M). Inference uses DDIM with 50 denoising steps to produce 16-frame clips at $512{\times}512$ resolution.
    
\item {\bf Code of ethics}
    \item[] Question: Does the research conducted in the paper conform, in every respect, with the NeurIPS Code of Ethics \url{https://neurips.cc/public/EthicsGuidelines}?
    \item[] Answer: \answerYes{}
    \item[] Justification: The research conforms with the NeurIPS Code of Ethics. The training data consists of publicly available YouTube videos (TalkVid) and the evaluation dataset (HDTF) is a standard benchmark in the portrait animation literature. The paper does not involve deception, human subjects experimentation, or collection of personally identifiable information beyond what is publicly available in existing datasets.

\item {\bf Broader impacts}
    \item[] Question: Does the paper discuss both potential positive societal impacts and negative societal impacts of the work performed?
    \item[] Answer: \answerYes{}
    \item[] Justification: Portrait animation technology has beneficial applications in dubbing, accessibility (enabling communication for individuals with facial paralysis or speech impairments), and virtual avatars, as noted in the introduction. However, the same technology can be misused for generating misleading video content (deepfakes). The parametric conditioning approach does not fundamentally alter this dual-use risk relative to existing portrait animation systems, but we acknowledge it. Detection and attribution methods developed by the broader community apply equally to Loki's outputs.
    
\item {\bf Safeguards}
    \item[] Question: Does the paper describe safeguards that have been put in place for responsible release of data or models that have a high risk for misuse (e.g., pre-trained language models, image generators, or scraped datasets)?
    \item[] Answer: \answerNA{}
    \item[] Justification: No models, code, or datasets are released at submission time. Upon acceptance, we plan to release code and weights with appropriate usage guidelines and terms of use that prohibit generation of non-consensual or deceptive content.

\item {\bf Licenses for existing assets}
    \item[] Question: Are the creators or original owners of assets (e.g., code, data, models), used in the paper, properly credited and are the license and terms of use explicitly mentioned and properly respected?
    \item[] Answer: \answerYes{}
    \item[] Justification: All datasets (TalkVid, HDTF), pretrained models (Stable Diffusion 2.1, FLAME, ArcFace), and baseline systems (X-Portrait, HunyuanPortrait, SadTalker, AniTalker, EchoMimic) are cited with their original publications. Stable Diffusion 2.1 is released under the CreativeML Open RAIL++-M License. FLAME is released under a non-commercial research license. All baselines are evaluated using their official publicly released code and checkpoints.

\item {\bf New assets}
    \item[] Question: Are new assets introduced in the paper well documented and is the documentation provided alongside the assets?
    \item[] Answer: \answerNA{}
    \item[] Justification: The paper does not release new datasets, models, or code at submission time. The two proposed evaluation metrics (HPF, HEF) are fully specified in \S\ref{sec:exp-metrics} with complete definitions and appendix derivations, enabling independent reimplementation.

\item {\bf Crowdsourcing and research with human subjects}
    \item[] Question: For crowdsourcing experiments and research with human subjects, does the paper include the full text of instructions given to participants and screenshots, if applicable, as well as details about compensation (if any)? 
    \item[] Answer: \answerNA{}
    \item[] Justification: The paper does not involve crowdsourcing or research with human subjects.

\item {\bf Institutional review board (IRB) approvals or equivalent for research with human subjects}
    \item[] Question: Does the paper describe potential risks incurred by study participants, whether such risks were disclosed to the subjects, and whether Institutional Review Board (IRB) approvals (or an equivalent approval/review based on the requirements of your country or institution) were obtained?
    \item[] Answer: \answerNA{}
    \item[] Justification: The paper does not involve research with human subjects.

\item {\bf Declaration of LLM usage}
    \item[] Question: Does the paper describe the usage of LLMs if it is an important, original, or non-standard component of the core methods in this research? Note that if the LLM is used only for writing, editing, or formatting purposes and does \emph{not} impact the core methodology, scientific rigor, or originality of the research, declaration is not required.
    \item[] Answer: \answerNA{}
    \item[] Justification: LLMs are not used as a component of the core method. Any use of LLMs was limited to writing assistance and does not impact the methodology, scientific rigor, or originality of the research.
\end{enumerate}

\end{document}